%% file: wregret.tex
\documentclass[12pt]{article}
\newcommand{\shortv}[1]{}
\newcommand{\fullv}[1]{#1}
\fullv{\input{defn.tex}}
\shortv{\input{siamdefn.tex}}
\shortv{\setcounter{secnumdepth}{3}}
\fullv{\usepackage{chicagor}}
\shortv{\newcommand{\citeyear}{\cite}}

\fullv{\input{spage.tex}}

\newcommand{\EU}{\mbox{EU}}
\renewcommand{\wr}{\mathit{wr}}
\newcommand{\ob}{\mathit{ob}}
\newcommand{\wc}{\mathit{wc}}
\newcommand{\mm}{\mathit{mm}}
\newcommand{\om}{\{\!\{}
\newcommand{\cm}{\}\!\}}
\newcommand{\regret}{\mathit{reg}}

\begin{document}
\title{Weighted regret-based likelihood: a new approach to describing
uncertainty} 
\shortv{\author{Joseph Y. Halpern\thanks{Supported in part by NSF grants 
IIS-0812045, IIS-0911036, and CCF-1214844, by AFOSR grants
FA9550-08-1-0438, FA9550-09-1-0266, and FA9550-12-1-0040, and by ARO
grant W911NF-09-1-0281.}}  
\institute{Cornell University\\Ithaca, NY 14853, USA\\
\email{halpern@cs.cornell.edu}}
}
\fullv{\author{Joseph Y. Halpern%
\thanks{Supported in part by NSF grants 
IIS-0812045, IIS-0911036, and CCF-1214844, AFOSR grant
FA9550-08-1-0438 and FA9550-09-1-0266, and ARO grant W911NF-09-1-0281.
A preliminary version of this paper appears in the \emph{Proceedings of
the 12th European Conference on Symbolic and
Quantitative Approaches to Reasoning with Uncertainty (ECSQARU)}, 2013.}
\\
Cornell University\\
halpern@cs.cornell.edu
}
}
%\shortv{\author{ }}
%\date{August, 2011}
%\shortv{\nocopyright}
\maketitle

\begin{abstract}
%\shortv{\begin{quote}}
%A new representation of likelihood is proposed, based on the notion of
%regret, and is completely characterized.
%joe5
Recently, Halpern and Leung \citeyear{HL12} suggested representing
uncertainty by a \emph{weighted} set of probability measures, and
suggested a way of making decisions based on this representation of
%joe6
%uncertainty: \emph{maximizing weighted regret}.  Their paper leaves does
uncertainty: \emph{maximizing weighted regret}.  Their paper does
not answer an apparently simpler question: what it means, according to
this representation of uncertainty, for an event $E$ to be more likely
than an event $E'$.  
In this paper, a notion of comparative likelihood when
uncertainty is represented by a weighted set of probability measures is
defined.  It generalizes the ordering defined by probability (and by lower
probability) in a natural way; a generalization of upper probability can
also be defined.  A complete axiomatic characterization of this notion of
regret-based likelihood is given. 
%\shortv{\end{quote}}
\end{abstract}

\section{Introduction}

Recently, Samantha Leung and I \cite{HL12} suggested representing
uncertainty by a \emph{weighted} set of probability measures, and
suggested a way of making decisions based on this representation of
uncertainty: \emph{maximizing weighted regret}.  
%joe6
%However, one thing 
%they do not do is
However, we did not answer an apparently simpler question: given this
representation of uncertainty, what does it mean for an event $E$ to be
more likely than an event $E'$?    This is what I do in this paper.
To explain the issues, I start by reviewing the Halpern-Leung approach.

It has frequently been observed that there are many situations where an
agent's uncertainty is not adequately described by a single probability
measure.  
\fullv{Specifically, a single measure may not be adequate
for representing an agent's ignorance.}  For example, there seems to be a big
difference between a coin known to be fair and a coin whose bias an
agent does not know, yet if the agent were to use a single measure
to represent her uncertainty, in both of these cases it would seem that the
measure that assigns heads probability $1/2$ would be used.  

One approach \fullv{that has been suggested} for representing ignorance is to
use a set $\P$ of probability measures \cite{Hal31}.  
%joe4
That approach has the benefit of representing uncertainty in general, not by a
single number, but by a range of numbers.   
%joe4
This allows us to
distinguish the certainty that a coin is fair (in which
case the uncertainty of heads is represented by a single number, $1/2$)
from knowing only that the probability of heads could be anywhere
between, say, $1/3$ and $2/3$.

But this approach
also has its 
problems.  For example, consider an 
agent who believes that a coin may have a slight bias.  Thus, although
it is unlikely to be completely fair, it is close to fair.  How should
we represent this with a set of probability measures?  Suppose that the
agent is quite sure that the bias is between $1/3$ and $2/3$.  We could,
of course, take $\P$ to consist of all the measures that give heads
probability between $1/3$ and $2/3$.  But how does the agent know that the
possible biases are \emph{exactly} between $1/3$ and $2/3$.  Does she
%joe4
%not consider $2/3 + \epsilon$ possible for some small $\epsilon$.   And
not consider $2/3 + \epsilon$ possible for some small $\epsilon$?   And
even if she is confident that the bias is between $1/3$ and $2/3$, this 
representation cannot take into account the possibility that she views biases
closer to $1/2$ as more likely than biases further from $1/2$.

There is 
%joe4
also
a second well-known concern: learning.  Suppose that the agent initially
considers possible all the measures that gives heads probability between
$1/3$ and $2/3$.  She then starts tossing the coin, and sees that,
of the first 20 tosses, 12 are heads.  It seems that the agent should then
consider a bias of greater than $1/2$ more likely than a bias of less
than $1/2$.  But if we use the standard approach to updating with sets
of probability measures (see \cite{Hal31}), and condition each of the
measures on the observation, since the coin tosses are viewed as
independent, the agent will continue to believe that the probability of
the next coin toss is between $1/3$ and $2/3$.  The observation has no
impact as far as learning to predict better.  The set $\P$ stays the
same, no matter what observation is made.

There is a well-known solution to these problems: using a second-order
measure on these measures to express how likely the agent
considers each of them to be.  (See \cite{Good80} for a discussion of
this approach and further references.)
%joe5
For example, an agent can express the fact that the bias of a coin is
more likely to be close to $1/2$ than far from $1/2$.  In addition, 
the problem of learning can be dealt with by straightforward
conditioning.  
But this approach leads to other problems.  Essentially,
it seems that the ambiguity that an agent might feel about the outcome
of the coin 
toss seems to have disappeared.  For example, suppose that the agent has
no idea what the bias is.  The obvious second-order probability to use is
the uniform probability on possible biases.  While we cannot talk about
the probability that the coin is heads\fullv{ (there is a set of probabilities,
after all, not a single probability)}, the \emph{expected} probability of
heads is $1/2$.  Why should an agent that has no idea of the bias of the
coin know or believe that the expected probability of heads is $1/2$?
\fullv{
Of course, if one had to use a single probability measure to
%joe6
%describe uncertainty, symmetry considerations dictate thta it should be
describe uncertainty, symmetry considerations dictate that it should be
the one that ascribes equal likelihood to heads and tails; similarly, if
one had to put a single second-order probability on the set of possible
biases, uniform probability seems like the most obvious choice.}  
%joe6
Moreover, if our interest is in making decisions, then maximizing the
expected utility using the expected probability again does not take the
agent's ignorance into account.
Kyburg \citeyear{Kyburg} and Pearl \citeyear{Pearl87} have even
argued that there is no need for a second-order probability on
probabilities; whatever can be done with 
a second-order probability can already be done with a basic probability.

Nevertheless, 
%joe5
%\fullv{In recent work \cite{HL11}, Leung and I proposed putting weights on
%measures.}
%\shortv{Recent work [ANON]%
%\footnote{The exact reference is omitted here to preserve anonymity; the
%paper is referred to as [ANON] throughout.}
%proposed putting weights on measures.}
%joe6
%But, when it comes to decisoin making,  it may be more reasonable
when it comes to decision-making,  it does seem useful 
to use an approach that represents ambiguity, while still
%joe6
%maintaining some of the useful features of having a second-order
maintaining some of the features of having a second-order
probability on probabilities.  One suggestion, made by 
Walley \citeyear{Walley97}, 
is to put a second-order possibility
measure on probability \shortv{measures; see also
\cite{ChateauneufFaro2009,Cooman05}.} 
%\shortv{measures.}
Leung and I similarly suggested putting weights on
each probability measure in $\P$.   Since we assumed that the weights
are normalized so that the supremum of the weights is 1, these weights
can also be viewed as a possibility measure.
%\footnote{We were unaware of Walley's work when we made our suggestion.}
If the set $\P$ is finite, we can also normalize so as to view the
weights as being second-order probabilities. 
%This approach is somewhat close to the idea of using probabilities
%(indeed, if the set $\P$ of probability measures is finite, then these
%weights can be taken to be probabilities), 
%but it has some advantages.  
%and indeed, has many of the advantages of a second-order probability,
%joe6
%while still allowing us to represent ambiguity.  Specfically,
%while still allowing us to represent ambiguity.  Specifically,
\fullv{As with second-order probabilities, the weights can vary over time, as
more information is acquired.  For example, we can start with a
state of complete ignorance (modeled by assuming that all probability
measures have weight 1), and update the weights after
making an observation $\ob$, we take the weight of a measure $\Pr$ to be the
relative likelihood of $\ob$ if $\Pr$ were the true measure.  (See
Section~\ref{sec:review} for details.)  With this approach, if there is a
true underlying measure generating the data, over time, the weight
of the true measure approaches 1, while the weight of all other
measures approaches 0.  Thus, this approach allows learning in a
natural way.  If, for example, the actual bias of the coin was $5/8$ in
the example above, no matter what the initial weights, as long as $5/8$
had positive weight, then its weight would almost surely converge to 1
as more observations were made, while the weight of all other
measures would approach 0.  
%joe5
This, of course, is exactly what would
happen if we had a second-order probability on $\P$.%
%This may not seem terribly surprising; the same would happen if we had a 
%measure over measures.  
The weights can also be used to represent the fact that some
probabilities in the set $\P$ are more likely than others.  
}
\shortv{As with second-order probabilities, the weights can vary over time, as
more information is acquired, and
can be used to represent the fact that some
probabilities in the set $\P$ are more likely than others.}  
%joe6
%The way this
%approach differs from using a second-order probability on $\P$.

%The innovation in the 
What makes this approach different from just using a second-order
probability on $\P$ lies
%\fullv{Halpern and Leung approach is in how decisions are made using these}
%\shortv{[ANON] approach is in how decisions are made using these}
%weighted measures.   
in how decisions are made.  
Leung and I used \emph{regret}, a standard approach
to decision-making that goes back to Niehans \citeyear{Niehans} and
Savage \citeyear{Savage51}.  If uncertainty is represented by a set $\P$
of probability measures, then regret works as
follows: for each
act $a$ and each measure $\Pr \in \P$, we can compute the
expected regret of $a$ with respect to $\Pr$; this is the difference
between the expected utility of 
$a$ and the expected utility of the act that gives the highest expected
utility with respect to $\Pr$.  We can then associate with an act $a$ 
its worst-case expected regret of $a$, over all measures $\Pr \in \P$,
and compare acts with respect to their worst-case expected regret.
With weights in the picture, we modify the
procedure by multiplying the expected regret associated with
measure $\Pr$ by the weight of $\Pr$, and compare acts according to their
worst-case \emph{weighted} expected regret.  
%joe6
This approach to making decisions is very different from that suggested
by Walley \citeyear{Walley97}.  Moreover,
using the weights in the way means that we cannot simply replace a
set of weighted probability measures by a single probability measure;
the objections of Kyburg \citeyear{Kyburg} and Pearl \citeyear{Pearl87} do not
apply.  

Leung and I \cite{HL12} show that this approach
seems to do reasonable things in a number of examples of interest, and
provide an elegant 
axiomatization of decision-making.  
\commentout{
However, one thing 
we did not do in our earlier paper is
to provide the answer to an apparently simpler question: given this
representation of uncertainty, what does it mean for an event $E$ to be
more likely than an event $E'$?}
So how can we represent relative likelihood using this approach?  
%joe6
This is something not considered in earlier papers using sets
of weighted probabilities.
%joe5
If uncertainty is represented by a single probability measure, the
answer is immediate: $E$ is more likely than $E'$
exactly if the probability of $E$ is greater than the probability of
$E'$.  
%joe5
When using sets of probability measures, various approaches have
been considered in the literature.  The most common takes $E$ to be more
likely than $E'$ if the \emph{lower probability} of $E$ is greater than
the lower probability of $E'$, where the lower probability of $E$ is its
worst-case probability, taken over the measures in $\P$ (see
Section~\ref{sec:relordering}).  We could also compare $E$ and $E'$ with
respect to their upper probabilities (the best-case probability with
respect to the measures in $\P$).  Another possibility is to take  $E$
to be more likely than $E'$ if $\Pr(E) \ge \Pr(E')$ for all measures
$\Pr \in \P$; this gives a partial order on likelihood. 

In this paper, I define a notion of relative likelihood when
uncertainty is represented by a weighted set of probability measures
that generalizes the ordering defined by lower probability 
in a natural way; I also define a generalization of upper probability.
We can then associate with an event $E$ two numbers that
are analogues of lower and upper probability.  If uncertainty is
represented by a single measure, then these two numbers coincide; in
general, they do not.  The interval can be thought of as representing
the degree of ambiguity in the likelihood of $E$.  Indeed, in the
special case when all the weights are 1, the numbers are essentially
just the lower and upper probability (technically, they are 1 minus the
lower and upper probability, respectively).
Interestingly, the approach to assigning likelihood is based on the
approach to decision-making.  Essentially, what I am doing is the
analogue of defining probability in terms of expected utility,
rather than the other way around.  
\commentout{
Indeed, the approach is really a
generalization of an approach for defining probability from expected
utility.}
The approach can be viewed as generalizing both probability
\shortv{and lower probability.}
%\footnote{If uncertainty is represented by a set
%$\P$ of probability measures, then the lower probability of an event $E$
%is its worst-case probability; see Section~\ref{sec:relordering}.} 
\fullv{and lower probability,
while at the same time allowing a natural approach to updating.}

Why we should be interested in such a
representation.  If all that we ever did with probability was to
use it to make decisions, then arguably this wouldn't be of much
interest; Halpern and Leung's work already shows how weighted sets of
probabilities can be used in decision-making.  The results of this paper
add nothing further to that question.  However, we often talk about the
likelihood of events quite independent of their use in decision-making
(think of the use of probability in physics, to take just one example).
Thus, having an analogue of probability seems important and useful in
its own right.

\fullv{
The rest of this paper is organized as follows.  
%\fullv{After reviewing the work of Halpern and Leung in}
%\shortv{After reviewing the work of [ANON] in}
After reviewing the relevant material in \cite{HL12} in
Section~\ref{sec:review}, I define regret-based likelihood in
Section~\ref{sec:relordering}, and compare it to lower probability.
I provide an axiomatic characterization of regret-based likelihood in
Section~\ref{sec:axiomatization}, 
and show how it relates to the axiomatic characterization of lower
probability.  
I conclude in Section~\ref{conclusion}.}

\section{Weighted Expected Regret: A Review}\label{sec:review}

Consider the standard setup in decision theory.  We have a state space
$S$ and an outcome space $O$.  An \emph{act} is a function from $S$ to
$O$; it describes an outcome for each state.  Suppose that we have a
utility function $u$ on outcomes and a set $\P^+$ of \emph{weighted 
probability measures}.  That is, $\P^+$ consists of pairs
$(\Pr,\alpha_{\Pr})$, where $\alpha_{\Pr}$ is a weight in $[0,1]$ and
$\Pr$ is a probability on $S$.  Let
$\P = \{\Pr: \exists \alpha( (\Pr,\alpha) \in \P^+)\}$.  
For each $\Pr \in \P$ there is assumed to be exactly one $\alpha$,
denoted $\alpha_{\Pr}$,  such that $(\Pr,\alpha) \in \P^+$.  
It is further assumed that weights have been normalized so that there is at
least one  
measure $\Pr \in \P$ such that $\alpha_{\Pr} = 1$.%
\footnote{The assumption that at least one probability measure has
%joe6
%a weight of 1 makes is convenient for comparison to other approaches;
a weight of 1 is convenient for comparison to other approaches;
see below.  
%joe6
However, making this assumption has no impact on the results of this
paper; as long as we restrict to sets where the weight is bounded, all
the results hold without change.
Note that the assumption that the weights are probabilities
runs into difficulties if we have an infinite number of measures in
$\P$; for example, if $\P$ includes all measures on heads from
$1/3$ to $2/3$, as discussed in the Introduction, using a uniform
probability, we would be forced to assign each individual probability
measure a weight of 0, which would not work well for our later
%joe3
%definitions.}  Finally, $\P^+$ is assumed to be closed, so that if
definitions.}  Finally, $\P^+$ is assumed to be \emph{weakly closed}, so
that if $(\Pr_n,\alpha_n) \in \Pr^+$ for $n=1, 2, 3, \ldots$, 
$(\Pr_n,\alpha_n) \rightarrow (\Pr,\alpha_{\Pr})$, and $\alpha_{\Pr} >
0$, then $(\Pr,\alpha_{\Pr})\in \P^+$.  (I discuss below why I require
$\P^+$ to be just weakly closed, rather than closed.) 

Where are the weights in $\P^+$ coming from?  In general, they can be
viewed them as subjective, just like the probability measures.  However,
as observed in \cite{HL12}, there is an important special case
where the weights can be given a natural interpretation.  Suppose that, as in
the case of the biased coin in the Introduction, we make observations in
a situation 
where the probability of making a given observation is determined by
some objective source.  Then we can start by giving all probability
measures a weight of 1.  Given an observation $\ob$ (e.g., sequence
of coin tosses in the example in the Introduction), we can compute
$\Pr(\ob)$ for each measure $\Pr \in \P$; we can then update the
weight of $\Pr$ to be $\Pr(\ob)/\sup_{{\Pr}' \in \P} {\Pr}'(\ob)$.  Thus,
the more likely the observation is according to $\Pr$, the higher the
updated weight of $\Pr$.%
\footnote{The idea of putting a possibility on probabilities in $\P$
that is determined by likelihood also appears in the work of Moral
\citeyear{Moral92}, although he does not consider a general approach to
dealing with sets of weighted probability measures.}
(The denominator is just a normalization to
ensure that some measure has weight 1.)   With this approach to
updating, 
if there is a true underlying measure generating the data, then
as an agent makes more observations, almost surely, the weight
of the true measure approaches 1, while the weight of all other
measures approaches 0.%
\footnote{The ``almost surely'' is due to the fact that, with
probability approaching 0, as more and more observations are made, it is
possible that an agent will make a misleading observations, that are not
representative of the true measure.  This also depends on the set of
possible observations being rich enough to allow the agent to ultimately
discover the true measure generating the observations.  Since learning
is not a focus of this paper, I do not make this notion of ``rich
enough'' precise here.}

%joe2:
I now review the definition of weighted regret, and introduce the notion
of \emph{absolute} (weighted) regret.  I start with regret.
The regret of an act $a$ in a state $s \in S$ is the difference between
the utility of the best act at state $s$ and the utility of $a$ at $s$.
%joe6
%Typically the comparison $a$ is not compared to all acts, but to the
Typically, the act $a$ is not compared to all acts, but to the
acts in a set $M$, called a \emph{menu}.  Thus, the regret of $a$ in
state $s$ relative to menu $M$, denoted 
$\regret^M(a,s)$, is \mbox{$\sup_{a' \in M} u(a'(s)) - u(a(s))$}.%
%joe6
\fullv{
\footnote{Recall that if $X$ is a set of real numbers, $\sup X$, the
\emph{supremum} of $X$, is the smallest real numbers that is greater than or
equal to all the elements of $X$.  If $X$ is finite, then the sup is the
same as the max.  But if $X$ is, say, the interval $(0,1)$, then $\sup X
= 1$.  Similarly, $\inf X$ is the largest real number that is less
than or equal to all the elements in $X$.}
}
There are typically some constraints put on $M$ to ensure that $\sup_{a'
\in M} u(a'(s))$ is finite---this is certainly the case if $M$ is finite,
or the convex closure of a finite set of acts, or if there is a best
possible outcome in the outcome space $O$.  The latter assumption 
holds in this paper, so I assume throughout that $\sup_{a'
\in M} u(a'(s))$ is finite.

For simplicity, I assume that the state space $S$ is finite.  
Given a probability measure $\Pr$ on $S$, the expected regret of
an act $a$ with respect to $\Pr$ relative to menu $M$ is just 
$\regret_{\Pr}^M(a) = \sum_{s \in S} \regret^M(a,s) \Pr(s)$.
The \emph{(expected) regret} of $a$ with respect to $\P$ and a
menu $M$ is
just the worst-case regret, that is, $$\regret_{\P}^M(a)
%= \sup_{\Pr \in \P} \alpha_{\Pr}(u(\bar{o}^*) - \EU_{\Pr}(a))$.
= \sup_{\Pr \in \P} \regret_{\Pr}^M(a).$$
%Note that, because the set $\{\alpha_{\Pr}(u(\bar{o}^*) - \EU_{\Pr}(a)):
%(\Pr,\alpha_{\Pr}) \in \P^+\}$ is bounded by $u(\bar{o}^*)$, the
%%supremum of this set exists and is at most $u(\bar{o}^*) +
%|u(\bar{o}_*)|$.  Moreover,
Similarly,
the \emph{weighted (expected) regret} of $a$ with respect to $\P^+$ and a
menu $M$ is
just the worst-case weighted regret, that is, $$\wr_{\P^+}^M(a)
%= \sup_{\Pr \in \P} \alpha_{\Pr}(u(\bar{o}^*) - \EU_{\Pr}(a))$.
= \sup_{\Pr \in \P} \alpha_{\Pr}\regret_{\Pr}^M(a).$$
%Note that, because the set $\{\alpha_{\Pr}(u(\bar{o}^*) - \EU_{\Pr}(a)):
%(\Pr,\alpha_{\Pr}) \in \P^+\}$ is bounded by $u(\bar{o}^*)$, the
%%supremum of this set exists and is at most $u(\bar{o}^*) +
%|u(\bar{o}_*)|$.  Moreover,
%joe6
%Thus, regret is a just a special case of weighted regret, where the
Thus, regret is just a special case of weighted regret, where 
%joe6
%the weights are all 1.  
all weights are 1.  

%joe3:
Note that, as far weighted regret goes, it does not hurt to augment a set
$\P^+$ of weighted probability measures by adding pairs of the form
$(\Pr,0)$ for $\Pr \notin \P$.  But if we start with an unweighted set
$\P$ of probability measures, the weighted set $\P^+ =
\{(\Pr,1): \Pr \in \P\} \union \{(\Pr,0): \Pr \notin \P\}$ is not
closed in general, although it is weakly closed. There may well be
a sequence $\Pr_n\rightarrow \Pr$, where $\Pr_n \notin \P$ for all $n$,
but $\Pr \in \P$.  But then we would have have $(\Pr_n,0) \in \P^+$
converging to $(\Pr,0) \notin \P^+$.  This is exactly why I required
only weak closedness.  Note for future reference that,
%joe3
%since $\P^+$ is assumed to be closed, there is some element
%$(\alpha_{\Pr},\Pr) \in 
since $\P^+$ is assumed to be weakly closed, if $\wr_{\P^+}^M(a) > 0$, then
there is some element
$(\Pr,\alpha_{\Pr}) \in 
\P^+$ such that $\wr_{\P^+}^M(a) = \alpha_{\Pr}\regret_{\Pr}^M(a)$.

Weighted regret induces an obvious preference order on acts: act $a$ is
at least as good as $a'$ with respect to $\P^+$ and $M$, written $a
\succeq_{\P^+,M}^{\regret} a'$, if 
%joe1
%$\wr_{\P}(a) \le  \wr_{\P}(a')$.  As usual, I write $a
$\wr_{\P^+}^M(a) \le  \wr_{\P^+}^M(a')$.  As usual, I write $a
\succ_{\P^+,M}^{\regret} a'$ if $a 
\succeq_{\P^+,M}^{\regret} a'$ but it is not the case that $a'
\succeq_{\P^+,M}^{\regret} a$.  The standard 
notion of regret is the special case of weighted regret where all
weights are 1.   
%\footnote{
%When considering regret with respect to a measure $\Pr$,
%it is  more standard to consider a menu $M$ of possible acts.  The regret of
%act $a$ relative to $M$ is the difference between the expected utility
%of $a$ and that of the best possible act in $M$.  This leads to an ordering
%$\succeq_{\P^+}^{\regret,M}$ on acts that is relative to $M$.  It is
%well known that we 
%can have $a \succ_{\P^+}^{\regret,M} a'$ and $a'
%\succ_{\P+}^{\regret,M'} a$, even if $M  
%\subseteq M'$, 
%and the only acts in $M' - M$ are acts that are ranked below both $a$
%and $a'$.  As long as there is a best-possible act ($\bar{o}^*$ in our
%case), and that act is in both $M$ and $M'$, we avoid this menu
%dependence.}  
I sometimes write $a \succeq_{\P,M}^{\regret} a'$ to denote the unweighted
case (i.e., where all the weights in $\P^+$ are 1).

In this setting, using weighted regret gives an approach that allows
an agent to transition smoothly from regret to expected utility. 
It is well known that regret generalizes 
expected utility in the sense that if $\P$ is a singleton $\{\Pr\}$,
then $\wr_{\P}^M(a) \le \wr_{\P}^M(a')$ iff $\EU_{\Pr}(a) \ge
\EU_{\Pr}(a')$ (where $\EU_{\Pr}(a)$ denotes the expected utility of act
%joe6
%$a$ with respect to probability $\Pr$).  (In praticular, this means that
$a$ with respect to probability $\Pr$).%
\footnote{This follows from the observation that, given a menu $M$,
there is a constant $c_M$ such that, for all acts $a \in M$,
$\wr^M_{\{\Pr\}}(a) = c_M - \EU_{\Pr}(a)$.} 
(In particular, this means that
if $\P$ is a singleton, regret is menu independent.)  
If we start with all the weights being 1, then, as observed above,
the weighted regret is just the standard notion of regret.  As the agent
makes observations, if there is a measure $\Pr$ generating the uncertainty,
the weights will get closer and closer to a
situation where $\Pr$ gets weight 1, with the weights of all
other measures dropping off quickly to 0, so the ordering of acts
will converge to the ordering given by expected utility with respect to $\Pr$.

There is another approach with some
similar properties, that again starts with uncertainty being represented
by a set $\P$ of (unweighted) probability measures.  Define 
$\wc_{\P}(a) = \inf_{\Pr \in \P} \EU_{\Pr}(a))$.  Thus $\wc_{\P}(a)$ is
the worst-case expected utility of $a$, taken over all $\Pr \in \P$. 
Then define $a \succeq_{\P}^{\mm} a'$ if $\wc_{\P}(a) \ge \wc_{\P}(a')$.
This is the maxmin expected utility rule, quite often used in economics
\cite{GS1989}.  There are difficulties in getting a weighted version of
maxmin expected utility \cite{HL12} (see
%\shortv{maxmin expected utility [ANON] (see}
Section~\ref{sec:relordering}); however, Epstein and Schneider 
\citeyear{epstein05a} propose another approach that can be combined with
maxmin expected utility.
They fix a parameter $\alpha \in (0,1)$, and update $\P$
after an observation $\ob$ by retaining only those measures $\Pr$
such that $\Pr(\ob) \ge \alpha$.  For any choice of $\alpha < 1$, we
again end up converging almost surely to a single measure, so again
this approach converges almost surely to expected utility.  

I conclude this section with a discussion of menu dependence.  Maxmin
expected utility is not menu dependent; the preference ordering on acts
induced by regret can be, as the following example illustrates.
\xam\label{xam:menu} Take the outcome space to be $\{0,1\}$, and the
utility function to be the identity, so that $u(1) = 1$ and $u(0) =
0$.  As usual, if $E \subseteq S$, $1_E$ denotes the
%joe6
%\emph{indicator function} on $E$, where, for for each state $s \in S$,
\emph{indicator function} on $E$, where, for each state $s \in S$,
we have $1_E(s) = 1$ if $s \in E$, and $1_E(s) = 0$ if $s \notin E$.
Let $S = \{s_1,s_2,s_3,s_4\}$, $E_1 = \{s_1\}$, $E_2 = \{s_2\}$, $E_3 =
\{s_2,s_3\}$, $M_1 = \{1_{E_1}, 1_{E_2}\}$,
$M_2 = \{1_{E_1}, 1_{E_2},1_{E_3}\}$, and $\P = \{\Pr_1, \Pr_2\}$, 
%joe6
%where $\Pr_1(s_1) = \Pr(s_3) = \Pr(s_4) = 1/3$, $\Pr_2(s_2) = 1/4$, and 
where $\Pr_1(s_1) = \Pr_1(s_3) = \Pr_1(s_4) = 1/3$, $\Pr_2(s_2) = 1/4$, and 
$\Pr_2(s_3) = 3/4$.
A straightforward calculation shows that 
$\regret_{\Pr_1}^{M_1}(1_{E_1}) = 0$,
$\regret_{\Pr_1}^{M_1}(1_{E_2}) = 1/3$,
$\regret_{\Pr_2}^{M_1}(1_{E_1}) = 1/4$,
$\regret_{\Pr_2}^{M_1}(1_{E_2}) = 0$,
$\regret_{\Pr_1}^{M_2}(1_{E_1}) = 1/3$,
$\regret_{\Pr_1}^{M_2}(1_{E_2}) = 2/3$,
$\regret_{\Pr_2}^{M_2}(1_{E_1}) = 1$, and
$\regret_{\Pr_2}^{M_2}(1_{E_2}) = 3/4$.
Thus, $1/4 = \regret^{M_1}_{\P}(1_{E_1}) < \regret^{M_1}_{\P}(1_{E_2}) =
1/3$, while 
$1= \regret^{M_2}_{\P}(1_{E_1}) > \regret^{M_2}_{\P}(1_{E_2}) = 3/4$.
The preference on $1_{E_1}$ and $1_{E_2}$ depends on whether we consider
%joe6
%$E_1$ or $E_2$.
the menu $M_1$ or the menu $M_2$. 
\exam

%An ``absolute'' notion of (weighted) regret, independent
%joe6
%of of the menu, can be defined if there is if there is an outcome $o^*
%\in O$  that give the maximum utility; that
%of the menu, can be defined if
Suppose that 
there is an outcome $o^*
\in O$  that gives the maximum utility; that
is, $u(o^*) \ge u(o)$ for all $o \in O$.  If
$\bar{o}^*$ is the constant act that gives outcomes $o^*$ in all
states, 
%it immediately follows that, for all acts $a$ and all measures $\Pr$, we
%have  $\EU_{\Pr}(a) \le \EU_{\Pr}(\bar{o}^*) = u(o^*)$.  That is, 
then $\bar{o}^*$ is clearly the best act in all states.
%Similarly, $\bar{o}_*$ is the worst possible act.
%As we shall see, in the context of interest for this paper, there is in
%fact a best act: $1_S$.
If there is such a best act, an ``absolute'', menu-independent notion
of weighted expected regret can be defined by always comparing to
$\bar{o}^*$.  That is,  define 
$$\begin{array}{ll}
\regret(s,a) = u(o^*) - u(a(s));\\
\regret_{\Pr}(a) = \sum_{s \in S} (u(o^*) - u(a(s))\Pr(s) = u(o^*) -
\EU_{\Pr}(a);\\
\fullv{\regret_{\P}(a) = \sup_{\Pr \in \P}\sum_{s \in S}(u(o^*) -
u(a(s))\Pr(s) = u(o^*) - \inf_{\Pr \in \P}(\EU_{\Pr}(a);\\}
\shortv{\regret_{\P}(a) = \sup_{\Pr \in \P}\sum_{s \in S}(u(o^*) -
u(a(s))\Pr(s)\\ \qquad\qquad = u(o^*) - \inf_{\Pr \in \P}(\EU_{\Pr}(a);\\
\wr_{\P^+}(a) = \sup_{\Pr \in \P} \alpha_{\Pr}\sum_{s \in S}(u(o^*) -
u(a(s))\Pr(s)\\ \qquad\qquad = \sup_{\Pr \in \P}\alpha_{\Pr} (u(o^*) - 
\EU_{\Pr}(a)).}
\fullv{\wr_{\P^+}(a) = \sup_{\Pr \in \P} \alpha_{\Pr}\sum_{s \in
S}(u(o^*) - u(a(s))\Pr(s) = \sup_{\Pr \in \P}\alpha_{\Pr} (u(o^*) - 
\EU_{\Pr}(a)).}
\end{array}
$$ 
If there is a best act, then I write $a \succeq_{\P^+} a'$ if
$\wr_{\P^+}(a) \le \wr_{\P^+}(a')$; similarly in the unweighted case, I
write $a \succeq_{\P} a'$ if $\wr_{\P}(a) \le \wr_{\P}(a')$.

Conceptually, we can think of the agent as always being aware of the
best outcome $o^*$, and comparing his actual utility with $a$ to
$u(o^*)$.  Equivalently, the absolute notion of regret is equivalent to
a menu-based notion with respect to a menu $M$ that includes $\bar{o}^*$
(since if the menu includes $\bar{o}^*$, it is the best act in every
state).  As we shall see, in our setting, we can always reduce 
menu-dependent regret to this absolute, menu-independent notion, since
there is in fact a best act: $1_S$.

\section{Relative Ordering of Events Using Weighted
Regret}\label{sec:relordering} 

%joe6
%Although the work of Halpern and Leung \citeyear{HL12} gives us a
%way of using weighted regret in decision-making, it does not deal with the
%arguably simpler problem of assigning relative likelihood to subset $E
%\subseteq S$.    Fortunately, there is a straightforward way of doing
%so.  
In this section, I consider how a notion of comparative likelihood can
be defined using sets of weighted probability measures.

As in Example~\ref{xam:menu}, take the outcome space to be $\{0,1\}$,
%joe6
%the utility function to be the indentity, and consider indicator
the utility function to be the identity, and consider indicator
functions.  It is easy 
to see that $\EU_{\Pr}(1_E) = \Pr(E)$, so that with this setup, we can
recover probability from expected utility.  Thus, if uncertainty is
represented by a single probability measure $\Pr$ and we make
%joe6
%decisions by prefering those acts that maximize expected utility, then
decisions by preferring those acts that maximize expected utility, then
we have $1_E \succeq 1_{E'}$ iff $\Pr(E) \ge \Pr(E')$.

Consider what happens if we apply this approach to maxmin expected utility.
%joe4
%Now we have that $1_E \succeq_{\Pr}^{\mm} 1_{E'}$ iff
Now we have that $1_E \succeq_{\P}^{\mm} 1_{E'}$ iff
$\inf_{\Pr \in \P} \Pr(E) \ge \inf_{\Pr \in \P} \Pr(E')$.
In the literature, $\inf_{\Pr \in \P} \Pr(E)$, denoted  $\P_*(E)$, is
called the \emph{lower probability} of $E$, and is a standard approach
to describing likelihood.  The dual \emph{upper probability}, $\sup_{\Pr
\in \P} \Pr(E)$, is denoted $\P^*(E)$. 
%joe4
An easy calculation shows that $\P^*(E) = 1 - \P_*(\overline{E})$
(where, as usual, $\overline{E}$ denotes the complement of $E$).
The interval $[\P_*(E),\P^*(E)]$ can be thought of as describing the
uncertainty of $E$; the larger the interval, the greater the ambiguity.

What happens if we apply this approach to regret?  First consider
unweighted regret.  If we restrict to acts of the form $1_E$, then the
best act is clearly $1_S$, which is just the constant function $1$.
Thus, we can (and do) use the absolute notion of regret here, and for
the remainder of this paper.
%joe6
%We then get that $1_E \succeq_{\Pr}^{\regret} 1_{E'}$ iff $\sup_{\Pr \in
We then get that $1_E \succeq_{\P}^{\regret} 1_{E'}$ iff $\sup_{\Pr \in
\P}(1 - 
\Pr(E)) \le  \sup_{\Pr \in \P}(1 - \Pr(E'))$ iff $\sup_{\Pr \in \P}
\Pr(\overline{E}) \le \sup_{\Pr \in \P} \Pr(\overline{E}')$; that is,
$\P^*(\overline{E}) \le \P^*(\overline{E}')$.
Moreover, easy manipulation shows that $\sup_{\Pr \in \P}(1 - \Pr(E)) = 1 -
\inf_{\Pr \in \P} \Pr(E) = 1 - \P_{*}(E)$.  It follows that $1_E
\succeq_{\P}^{\regret} 1_{E'}$ iff  $(1 - \P_*(E)) \le   (1-\P_*(E'))$ iff
%joe4
$\P_*(E) \ge \P_*(E')$ iff $1_E \succeq_{\P}^{\mm} 1_{E'}$;
both regret
and maxmin expected utility put the same ordering on events.

The extension to weighted regret is immediate.  Let $\P^+_{\regret}(E)$,
the \emph{(weighted) regret-based likelihood of $E$}, be defined as
$\sup_{\Pr \in\P} \alpha_{\Pr} \Pr(\overline{E})$.   If $\P^+$ is
unweighted, so that all the weights are 1, I write $\P_{\regret}(E)$ to
denote $\sup_{\Pr \in\P} \Pr(\overline{E})$.
Note that $\P_{\regret}(E) = 1 - \P_*(E)$, so $\P_{\regret}(E) \le
\P_{\regret}(E')$ iff  
%joe6
%$\P_*(E) \ge \P_I(E')$.  That is, the ordering induced by $\P_{\regret}$
$\P_*(E) \ge \P_*(E')$.  That is, the ordering induced by $\P_{\regret}$
is the opposite of that induced by $\P_*$.  So, for example,
$\P_{\regret}(\emptyset) = 1$ and  $\P_{\regret}(S) = 0$; smaller sets
have a larger regret-based likelihood.%
\footnote{Since an act with smaller regret is viewed as better, the
ordering on acts of the form $1_E$ induced by regret is the same as that
induced by maxmin expected utility.}

Regret-based likelihood provides a way of associating a number with each
event, just as probability and lower probability do. 
Moreover, just as lower probability gives a lower bound on uncertainty,
we can think of $\P_{\regret}^+(E)$ as giving an upper bound on the
uncertainty.  (It is an upper bound rather than a lower bound because
larger regret means less likely, just as smaller lower probability does.)
The naive corresponding lower bound is given by $\inf_{\Pr \in\P} \alpha_{\Pr}
\Pr(\overline{E})$.  This lower bound is 
not terribly interesting; if there are probability measures $\Pr' \in
\P$ such that $\alpha_{\Pr'}$ is close to 0, then this lower bound will
be close to 0, independent of the agent's actual feeling about the
likelihood of $E$.  
%Since we expect that, with learning, the weight of the
%``true'' measures will approach 1, while the weight of the remaining
%measures will approach 0, this says that, in many cases of
%interest, $\inf_{\Pr \in\P} \alpha_{\Pr} \Pr(\overline{E})$ will be
%very close to 0, which is not informative.  
%
A more reasonable lower bound is given by the
expression $\underline{\P}_{\regret}^+(E) = 1 -
\P^+_{\regret}(\overline{E})$ (recall that the analogous expression
relates upper probability and lower probability).  The intuition for
this choice is the 
following.  If nature were conspiring against us, she would try to prove
us wrong by making $\alpha_{\Pr} \Pr(\overline{E})$ as large as
%joe6
%possible---that is, make the weighted probablity of being wrong a large
possible---that is, make the weighted probability of being wrong as large
as possible. On the other hand, if nature were conspiring with us, she
would try to make $\alpha_{\Pr}\Pr(E)$ as large as possible,
or, equivalently, make $1-\alpha_{\Pr}\Pr(E)$ as small as possible.
Note that
this is different from making $\alpha_{\Pr}\Pr(\overline{E})$ as large
as possible, unless $\alpha_{\Pr} = 1$ for all $\Pr \in \P$.  An
easy calculation shows that 
$$\begin{array}{lll}
1 - {\P}^+_{\regret}(\overline{E}) &= &1-\sup_{\Pr \in \P}
\alpha_{\Pr}\Pr(E) \\
&= &\inf_{\Pr \in \P} (1 - \alpha_{\Pr} \Pr(E)).
\end{array}
$$
This motivates the definition of $\underline{\P}^+_{\regret}$.

The following lemma clarifies the relationship between these
expressions, and shows that
$[\underline{\P}_{\regret}^+(E),\P_{\regret}^+(E)]$ really does give an
interval of ambiguity.  
%\shortv{(The proofs of all technical results can
%be found in the full paper, which is submitted as supplementary material.)}

\lem\label{lem:dual} $\inf_{\Pr \in \P}\alpha_{\Pr}\Pr(\overline{E})
\le 1 - {\P}^+_{\regret}(\overline{E}) \le {\P}^+_{\regret}(E)$.%
\shortv{\footnote{The proof of this result and all others can be found
in the full paper, available at
http://www.cs.cornell.edu/home/halpern/papers/wregret.pdf.}} 
\elem

\fullv{
\prf Clearly 
%$$\begin{array}{lll}
$$\inf_{\Pr \in\P} \alpha_{\Pr} \Pr(\overline{E}) = \inf_{\Pr \in
\P} \alpha_{\Pr} (1 - \Pr(E)).$$
%&= &\sup_{\Pr \in \P}(1 - \alpha_{\Pr} + \alpha_{\Pr} \Pr(E)).
%\end{array}
%$$
Since, as observed above, $$1 - {\P}^+_{\regret}(\overline{E}) = \inf_{\Pr
\in \P} (1 - \alpha_{\Pr} \Pr(E)),$$ and  
for all $\Pr \in\P$, we have $$1-\alpha_{\Pr} \Pr(E) 
\ge \alpha_{\Pr}(1-\Pr(E)),$$ 
it follows that $\inf_{\Pr \in \P}\alpha_{\Pr}\Pr(\overline{E}) \le 1 -
{\P}^+_{\regret}(\overline{E})$.  

Since, by assumption, there is a probability measure ${\Pr}' \in
\P$ such that $\alpha_{{\Pr}'} = 1$, it follows that
$$\begin{array}{lll}
1 - {\P}^+_{\regret}(\overline{E}) &= &1 -\sup_{\Pr\in\P} \alpha_{\Pr}
\Pr(E)\\ &\le &1 - {\Pr}'(E) = {\Pr}'(\overline{E})\\  &\le 
&\sup_{\Pr \in \P}\alpha_{\Pr} \Pr(\overline{E})\\
&\le &\P^+_{\regret}(E).
\end{array}$$ 
\eprf
}

In general, equality does not hold in Lemma~\ref{lem:dual}, as shown by
the following example.  The example also illustrates how the ``ambiguity
%joe6
%interval'' can decreases with weighted regret, if the weights are
interval'' can decrease with weighted regret, if the weights are
updated as suggested in \cite{HL12}.

\xam Suppose that the state space consists of $\{h,t\}$ (for heads and
tails); let $\Pr_{\beta}$ be the measure that puts probability $\beta$
on $h$.  Let $\P^+_0 = \{(\Pr_{\beta},1): 1/3 \le \beta \le 2/3\}$.
That is, we initially consider all the measures that put
probability between $1/3$ and $2/3$ on heads.  We toss the coin and
observe it land heads.  Intuitively, we should now consider it more
likely that the probability of heads is greater than $1/2$.  Indeed,
applying likelihood updating, we get the set $\P_1^+ = \{(\Pr_{\beta},
3\beta/2): 1/3 \le \beta \le 2/3\}$;%  
\footnote{The weight of $\Pr_{\beta}$ is the likelihood of
observing heads according to $\Pr_{\beta}$, which is just $\beta$,
normalized by the likelihood of observing heads according to the measure
that gives heads the highest probability, namely $2/3$.}  
the probability measures that give $h$ higher probability get higher weight.
In particular, the weight of $\Pr_{2/3}$ is
still 1, but the weight of $\Pr_{1/3}$ is only $1/2$.
If the coin is
tossed again and this time tails is observed, we update
further to get $\P_2^+ = \{(\Pr_{\beta}, 4\beta(1-\beta)): 1/3 \le \beta \le
2/3\}$.  An 
easy calculation shows 
that $[\underline{\P}_{0,\regret}^+(h),{\P}_{0,\regret}^+(h)] = [1/3,2/3]$, 
$[\underline{\P}_{1,regret}^+(h),\P_{1,\regret}^+(h)] = [1/3,3/8]$, and 
$[\underline{\P}_{2,\regret}^+(h),\P_{2,\regret}^+(h)] = [11/27,16/27]$.  

It is also
easy to see that $\inf_{\Pr} 4\beta(1-\beta) \Pr_{\beta}(t) = 8/27$, so
$\inf_{\Pr \in \P_2} 4\beta(1-\beta)\Pr_{\beta}(t) < 1 -
\P^+_{2,\regret}(t) < \P^+_{2,\regret}(h)$.
Thus, for $\P^+_2$, we get strict inequalities for the
expressions in Lemma~\ref{lem:dual}.
\exam

%joe1*
\commentout{
Interestingly, there does not seem to be an obvious way to define a
weighted version of lower probability.  The obvious
definition,$\inf_{\Pr \in \P} 
\alpha_{\Pr} \Pr(E)$, does not seem to capture reasonable intuitions.  
For example, going back to the example of the biased coin from the
Introduction, suppose that $\Pr$ consists of only two measures, 
the first, which gets weight 1, gives heads probability $2/3$; the
second, which gets weight $\epsilon$, gives heads probability $1/3$.
Thus, the agent is almost certain that heads should get probability
$2/3$.  Nevertheless, if $H$ is the event of heads on the next coin toss,
$\inf_{\Pr \in \P} \Pr(H) = 1/3\epsilon$.  By way of contrast, weighted
upper probability seems to be a better-behaved notion.  
%Lower probability can be thought of as a conservative, or pessimistic
%notion of assigning likelihood, as is regret; upper probablity is an
%optimistic notion.  
}

The width of the interval
$[\underline{\P}^+_{\regret}(E), \P^+_{\regret}(E)]$ can be viewed as a
measure of the ambiguity the agent feels about $E$, just as the interval
$[\P_*(E), \P^*(E)]$.  Indeed, if all the weights are 1, the two
intervals have the same width, since $\P_*(E) = 1 - \P^+_{\regret}(E)$
and $\P^*(E) = 1 - \underline{\P}^+_{\regret}(E)$ in this case.

However, weighted regret has a significant
advantage over upper and lower probability here.  If the true bias of
the coin is, say $5/8$, then if the set $\P_k^+$ represents the
uncertainty after $k$ steps, as $k$ increases, almost surely,
$[\underline{\P}_{k,\regret}^+(h),\P_{k,\regret}^+(h)]$ will be a  
smaller and smaller interval containing $1-5/8=3/8$.  More
generally, using likelihood updated combined with weighted regret
provides a natural way to model the reduction of ambiguity via learning.

One concern with the use of regret has been the dependence of regret on
the menu.  
It is also worth noting that, in this context, there is a sense in which
we can work with the absolute notion of weighted regret without loss of
generality:  if we restrict to indicator functions,  
%joe1*
then a preference relative to a menu can
always be reduced to an absolute preference.
Given a
menu $M$ consisting of indicator functions, let $E_M = \union\{E: 1_E
\in M\}$.  
%Given $E' \subseteq S$, define the mneu  $M + 1_{E'} = \{1_{E
%\union E'}: 1_E \in M\}$;
%joe6
that is, $E_M$ is the union of the events for which the corresponding
indicator function is in $M$.  
%In the following result, as usual, I use
%$\overline{E}$ to denote the complement of a set $E$.

\pro If $M$ is a menu consisting of indicator functions, 
%joe4
%such that 
and
$1_{E_1}, 1_{E_2} \in M$, then $1_{E_1} \succeq_{\P^+,M}^{\regret}
1_{E_2}$ iff $1_{E_1} + 1_{\overline{E}_M} \succeq^{\regret}_{\P^+} 1_{E_2}
+ 1_{\overline{E}_M}$. \epro

\fullv{
\prf Let $M'$ be any menu consisting of indicator functions that
includes $1_{E_1} + 1_{\overline{E}_M}$, 
$1_{E_2} + 1_{\overline{E}_M}$, and $1_S$.  Recall that 
$1_{E_1} + 1_{\overline{E}_M} \succeq^{\regret}_{\P^+} 1_{E_2}
+1_{\overline{E}_M}$ iff  
$1_{E_1} + 1_{\overline{E}_M} \succeq^{\regret}_{M',\P^+}
1_{E_2} + 1_{\overline{E}_M}$; the absolute notion of regret is
equivalent to the menu-based notion, as long as the menu includes the
best act, which in this case is $1_S$.  It clearly suffices to show
that, for all states $s \in S$ and all acts $1_{E} \in M$, 
$$\regret^M(1_{E},s) = \regret^{M'}(1_{E} + 1_{\overline{E}_M},s).$$
This is straightforward.  There are two cases, depending on whether $s
\in E_M$.

If $s \in E_M$, then, by definition, there is some act $1_{E'} \in M$
such that $s \in E'$, so $\sup_{a \in M} u(a(s)) = u(1)$.  Clearly
$\sup_{a \in M'} u(a(s)) = u(1)$, since $1_S \in M'$.  Moreover,
$1_{\overline{E}_M}(s) = 0$, so $(1_E + 1_{\overline{E}_M})(s) = 
1_E(s)$.  Thus, for $s \in E_M$,
$$\begin{array}{ll}
\regret^M(1_E,s) &= \sup_{a \in M} u(a(s)) - u(1_E(s)) \\ &= 
\sup_{a \in M'} u(a(s)) - u((1_E + 1_{\overline{E}_M})(s))\\ 
&= \regret^{M'}(1_E + 1_{\overline{E}_M},s).\end{array}$$
For $s \notin \overline{E}_M$, we have $a(s) = 0$ for all $a \in M$ and 
$1_E(s) = 0$, so $\sup_{a \in M} u(a(s)) - u(1_E(s)) = 0$.  
On the  other hand, $\sup_{a \in M'} u(a(s)) = u(1)$, and 
$u((1_E + 1_{\overline{E}_M})(s)) = u(1)$, so again
$\sup_{a \in M'} u(a(s)) - u((1_E + 1_{\overline{E}_M})(s)) = 0$.
Thus, we again have
$\regret^M(1_E,s) = \regret^{M'}(1_E+1_{\overline{E}_M},s)$.
\eprf
}

\section{Characterizing Weighted Regret-Based Likelihood}\label{sec:axiomatization} 

The goal of this section is to characterize weighted regret-based likelihood
axiomatically.  In order to do so, it is helpful to review the
characterizations of probability and lower probability.

A probability measure on a finite set $S$ maps subsets of $S$ to $[0,1]$
in a way that satisfies the following three properties:%
\footnote{Since I assume that $S$ is finite here, I assume that all
probability measures have domain $2^S$, and ignore measurability issues.}
%\begin{itemize}
\begin{description}
\item[Pr1.] $\Pr(S) = 1$.
\item[Pr2.] $\Pr(\emptyset) = 0$.%
\footnote{This property actually follows from the other two, using the
observation that $\Pr(S \union \emptyset) = \Pr(S) + \Pr(\emptyset)$; I
%joe6
%include it here to make the comparison to other approaches easier.}
include it here to ease the comparison to other approaches.}
\item[Pr3.] $\Pr(E \union E') = \Pr(E) + \Pr(E')$ if $E \inter E' =
\emptyset$.%
%\end{itemize}
\end{description}
These three properties characterize probability in the sense that any
function $f: 2^S \rightarrow [0,1]$ that satisfies these properties is a
probability measure.

Lower probabilities satisfy analogues of these properties:
%\begin{itemize}
\begin{description}
\item[LP1.] $\P_*(S) = 1$.
\item[LP2.] $\P_*(\emptyset) = 0$.
\item[LP3$'$.] $\P_*(E \union E') \ge \P_*(E) + \P_*(E')$ if $E \inter E' =
\emptyset$.
%\end{itemize}
\end{description}
However, these properties do not characterize lower probability.  There
are functions that satisfy LP1, LP2, and LP3$'$ that are not the lower
probability corresponding to some set of probability measures. (See
\cite[Proposition 2.2]{HalPuc00} for an example showing that analogous
properties do not characterize $\P^*$; the same example also shows that
they do not characterize $\P_*$.)

Various characterizations of $\P_*$ (and $\P^*$) have been proposed in
the literature
\fullv{\cite{Anger85,Giles82,Huber76,Huber81,Lorentz52,Williams76,Wolf77},}
\shortv{\cite{Anger85,Huber76,Huber81,Lorntz52,Williams76,Wolf77},}
all
similar in spirit.  I discuss one due to Anger and Lembcke
\citeyear{Anger85} here, since it makes the contrast between lower
probability and regret particularly clear.
The characterization is based on the notion of
{\em set cover}: a set $E$ is said to be covered
%joe7
%$n$ times by a multiset $M$ of sets if every
$n$ times by a multiset $M$ if every
element of $E$ appears at least $n$ times in $M$. 
It is important to note here that $M$ is a multiset,
not a set; its elements are not necessarily distinct.
(Of course, a set is a special case of a multiset.)
Let $\sqcup$ denote multiset union; thus, if $M_1$ and $M_2$ are
multisets, then $M_1 \sqcup M_2$ consists of all the elements in $M_1$
or $M_2$, which appear with multiplicity that is the sum of the
multiplicities in $M_1$ and $M_2$.  For example, using the $\om \ldots
\cm$ notation to denote a multiset, then $\om 1, 1, 2\cm \sqcup \om 1,
2, 3 \cm = \om 1, 1, 1, 2, 2, 3 \cm$.  

If $E \subseteq S$, then an {\em $(n,k)$-cover of $(E,S)$\/} is a multiset $M$
that covers $S$ $k$ times and covers $E$ $n+k$ times.  Multiset $M$
is an \emph{n-cover} of $E$ if $M$ covers $E$ $n$ times.
%joe4
For example, if $S = \{1,2,3\}$, then $\om 1,1,1,2,2,3\cm$ is a
$(2,1)$-cover of $(\{1\},S)$, a $(1,1)$-cover of $(\{1,2\},S)$, and a
3-cover of $\{1\}$.
Consider the following property:
%\begin{itemize}
\begin{description}
\item[LP3.] For all integers $m,n,k$ and all subsets
$E_1, \ldots, E_m$ of $S$, if $ E_1 \sqcup \ldots \sqcup E_m$ is an
$(n,k)$-cover of $(E,S)$, then $k+n\P_*(E)\geq\sum_{i=1}^{m}\P_*(E_i)$.%
\footnote{Note that LP3 implies LP2, using the fact that $\emptyset
\sqcup \emptyset$ is a (1,0)-cover of $(\emptyset,S)$.}
%\end{itemize}
\end{description}
There is an analogous property for upper probability, where $\ge$ is
replaced by $\le$.
It is easy to see that LP3 implies LP3$'$ (since $E \sqcup E'$ is a
$(1,0)$ cover of $E \union F$).

\thm {\rm \cite{Anger85}}\label{t:upm}
If $f:2^S \rightarrow[0,1]$, then there
exists a set $\P$ of probability measures with $f=\P_*$ if and
only if $f$ satisfies LP1, LP2, and LP3.  
%Moreover, if $f$ satisfies
%LP1, LP2, and LP3, then there is a unique closed convex set $\P$ of
%probability measures with $f=\P_*$.  
\ethm

Moving to regret-based likelihood, clearly we have 
%\begin{itemize}
\begin{description}
\item[REG1.] $\P^+_{\regret}(S) = 0$.
\item[REG2.] $\P^+_{\regret}(\emptyset) = 1$.
%\end{itemize}
\end{description}
The whole space $S$ has the least
regret; the empty set has the greatest regret.   In the unweighted case,
since $\P_{\regret}(E) = \P^*(\overline{E})$, REG1, REG2, and the
following analogue of LP3 (appropriately modified for $\P^*$) clearly
characterize $\P_{\regret}$:
%\begin{itemize}
\begin{description}
\item[REG3$'$.] For all integers $m,n,k$ and all subsets
$E_1, \ldots, E_m$ of $S$, if $ \overline{E}_1
\sqcup \ldots \sqcup \overline{E}_m$ is an 
$(n,k)$-cover of $(\overline{E},S)$, then
$k+n\P_{\regret}(E)\le \sum_{i=1}^{m}\P_{\regret}(E_i)$. 
%\end{itemize}
\end{description}
Note that complements of sets ($\overline{E}_1, \ldots, \overline{E}_m,
\overline{E}$) are used here, since regret is minimized if the
probability of the complement is maximized.  This need to work with the
%joe1
%complement makes the statement of the properties (and the proofs below)
complement makes the statement of the properties (and the proofs of the
theorems) 
slightly less elegant, but seems necessary.

%joe6
%It is not hard to see that REG3$'$ does not holds for weighted
It is not hard to see that REG3$'$ does not hold for weighted
regret-based likelihood.  For example, suppose that $S = \{a,b,c\}$ and
$\P^+ = ((\Pr_1, 2/3), (\Pr_2, 2/3), (\Pr_3, 1))$, where, 
identifying the probability $\Pr$ with the tuple $(\Pr(a), \Pr(b),
\Pr(c))$, we have 
\fullv{
\begin{itemize}
\item $\Pr_1 = (2/3, 0, 1/3)$;
\item $\Pr_2 = (1/3, 0, 2/3)$;
\item $\Pr_3 = (1/3, 1/3, 1/3)$.
\end{itemize}
}
\shortv{$\Pr_1 = (2/3, 0, 1/3)$, $\Pr_2 = (1/3, 0, 2/3)$, and
$\Pr_3 = (1/3, 1/3, 1/3)$.}
Then 
$\P_{\regret}^+(\{a,b\}) = \P_{\regret}^+(\{b,c\}) = 4/9$, while 
$\P_{\regret}^+(\{b\}) = 2/3$.  Since $\{a,b\} \sqcup \{b,c\}$ is a
(1,1)-cover of $(\{b\},\{a,b,c\})$, REG3$'$ would require that  
$$\P_{\regret}^+(\{a,b\}) + \P_{\regret}^+(\{b,c\}) \ge 1 +
\P_{\regret}^+(\{b\}),$$
which is clearly not the case.

We must thus weaken REG3$'$ to capture weighted regret-based
likelihood.  It turns out that the appropriate weakening is the
following:
%\begin{itemize}
\begin{description}
\item[REG3.] For all integers $m,n$ and all subsets
$E_1, \ldots, E_m$ of $S$, if $ \overline{E}_1
\sqcup \ldots \sqcup \overline{E}_m$ is an 
$n$-cover of $\overline{E}$, then
$n\P^+_{\regret}(E)\le \sum_{i=1}^{m}\P^+_{\regret}(E_i)$.%
%\end{itemize}
\end{description}

Although REG3 is weaker than REG3$'$, it still has some nontrivial
consequences.  For example, it follows from REG3 that $\P^+_{\regret}$ is
anti-monotonic.  If $E \subseteq E'$, then $\overline{E}$ is a 1-cover
of $\overline{E}'$, so by REG3, we must have $\P^+_{\regret}(E) \ge
\P^+_{\regret}(E')$.   Since $E \sqcup E'$ is trivially a 1-cover of $E
\union E'$, it also follows that $\P^+_{\regret}(\overline{E}) +
\P^+_{\regret}(\overline{E}') \ge \P^+_{\regret}(\overline{E \union E'})$.
REG3 also implies REG1, since $\emptyset$ (= $\overline{S}$) is an
$n$-cover of itself for all $n$.  

I can now state the representation theorem.  It says that a
representation of uncertainty satisfies REG1, REG2, and REG3 iff it is
the weighted regret-based likelihood determined by some set $\P^+$.  The set 
$\P^+$ is not unique, but it can be taken to be \emph{maximal}, in the
sense that if weighted regret-based likelihood with respect to some
other set 
$(\P')^+$ gives the same representation, then for all pairs
$(\Pr,\alpha') \in (\P')^+$, there exists $\alpha \ge \alpha'$ such
that $(\Pr,\alpha) \in \P^+$.  This (unique) maximal set $\P^+$ 
can be viewed as the canonical representation of uncertainty.
%\fullv{(See below for further discussion of uniqueness.)}

\thm \label{thm:regretchar}
If $f:2^S \rightarrow[0,1]$, then there
%joe3
%exists a closed set $\P^+$ of weighted probability measures with
exists a weakly closed set $\P^+$ of weighted probability measures with
$f=\P^+_\regret$ if and only if $f$ satisfies REG1, REG2, and REG3;
moreover, $\P^+$ can be taken to be maximal.
\ethm

%joe3
%\prf Clearly, given a closed set $\P^+$ of weighted probability
\fullv{
\prf Clearly, given a weakly closed set $\P^+$ of weighted probability
measures, the function $\P^+_\regret$ satisfies  REG1 and REG2.  To see
that it 
satisfies REG3, 
%joe3: 
%$\overline{E}_1, \ldots, \overline{E}_m$ of $S$, 
suppose that $\overline{E}_1 \sqcup \ldots \sqcup \overline{E}_m$ is an 
$n$-cover of $\overline{E}$.  
If $\P^+_{\regret}(E) = 0$, then REG3 trivially holds.  
If $\P^+_{\regret}(E) > 0$, then 
%joe3
%since $\P^+$ is closed,  
since $\P^+$ is weakly closed,  
there must be some probability $\Pr \in \P$ such that $\P^+_{\regret}(E) =
\alpha_{\Pr} \Pr(\overline{E})$.  Since 
$\overline{E}_1 \sqcup \ldots \sqcup \overline{E}_m$ is an 
$n$-cover of $\overline{E}$, it is easy to see that 
$\Pr(\overline{E}_1) + \cdots + \Pr(\overline{E}_m) = n
 \Pr(\overline{E})$, so 
$\alpha_{\Pr} \Pr(\overline{E}_1) + \cdots + \alpha_{\Pr}
\Pr(\overline{E}_m) = n \alpha_{\Pr} \Pr(\overline{E})$.
But $\alpha_{\Pr} \Pr(\overline{E}) = \P^+_{\regret}(E)$, by construction,
and $\alpha_{\Pr} \Pr(\overline{E}_i) \le \P^+_{\regret}(E_i)$, $i=1,
\ldots, n$.  Thus, 
$n\P^+_{\regret}(E)\le \sum_{i=1}^{m}\P^+_{\regret}(E_i)$. 

For the opposite direction, suppose that $f: 2^S \rightarrow [0,1]$
satisfies REG1, REG2, and REG3.  Let $\P = \Delta(S)$, the set of all
probability measures on $S$, and for $\Pr \in \P$, define
$$\alpha_{\Pr} = \sup \{\beta: \beta \Pr(\overline{E}) \le f (E) \mbox{
for all } E 
\subseteq S\}.$$
Note that, for all $\Pr \in \P$, we have $0 \Pr(\overline{E}) \le f(E)$ for
all $E \subseteq S$, since 
$f(E) \in [0,1]$, and $1 \Pr(\overline{\emptyset}) = f(\emptyset)
= 1$.
It follows that $\alpha_{\Pr} \in [0,1]$ for all $\Pr \in \P$.  Let $\P^+ =
\{(\Pr,\alpha_{\Pr}): \Pr \in \Delta(S)\}$.  It is easy to see that $\P^+$
is 
%joe3
weakly 
closed.  We want to show that there exists $\Pr \in \Delta(S)$ such
that (1) $\alpha_{\Pr} = 1$ (since this is one of the conditions on sets of
weighted measures) and (2) $f(E)= \P^+_{\regret}(E)$ for all $E
\subseteq S$.

The proof of this result makes critical use of the following variant of
Farkas' Lemma 
\cite{Farkas} (see \cite[page 89]{Schrijver}) 
from linear programming, where $A$ is a matrix, $b$ is a column vector,
and $x$ is a column vector of distinct variables:

\lem\label{farkas}
If $Ax \geq b$ is
unsatisfiable, then there exists a row vector
$\beta$ such that
\begin{enumerate}
\item
$\beta \geq 0$
\item
$\beta A = 0$
%(where $\alpha^T$ is the transpose of $\alpha$)
\item
$\beta b > 0$.
\end{enumerate}
\elem

\noindent Intuitively,
$\beta$ is a ``witness'' of the fact that $Ax \geq b$ is unsatisfiable.
This is because if there were a vector
$x$ satisfying $Ax \geq b$, then
$0 = (\beta A)x = \beta (Ax) \geq \beta b > 0$, a contradiction.

To prove the first claim, suppose that $S= \{s_1, \ldots, s_N\}$.  I
now construct a set of linear equations in the variables $x_1, \ldots,
x_N$ such that a solution to the equations guarantees the existence of a
probability measure $\Pr \in \Delta(S)$ such that $\alpha_{\Pr}= 1$.
Intuitively, we want $x_i$ to be $\Pr(x_i)$.  Since we must have
$\Pr(\overline{E}) \le f(E)$ for all $E \subset S$,%
\footnote{I use $\subset$ to denote strict subset.}
 for each $E \subset S$,
we have the inequality
$\sum_{\{i: s_i \notin E\}} x_i \le f(E)$.  Note that since 
$f(\emptyset) = 1$, the equation when $E = \emptyset$ is $x_1 + \cdots +
x_N \le 
1$.  In addition, we require that $x_i \ge 0$ for $i =1, \ldots, N$, and that 
$x_1 + \cdots + x_N = 1$.  It suffices to require that $x_1 + \cdots +
x_n \ge 1$, since, as I observed earlier, the equation corresponding to
$E = \emptyset$ already says $x_1 + \cdots + x_n \le 1$.
To apply Farkas' Lemma all the 
inequalities need to involve $\ge$, so this collection of inequalities
must be rewritten as:  
$$\begin{array}{l}
-\sum_{\{i: s_i \notin E\}} x_i \ge -f(E), \mbox{ for all
$E \subset S$}\\
x_i \ge 0, \mbox{ for $i = 1, \ldots, N$}\\
x_1 + \cdots + x_N \ge 1.
\end{array}$$
This system of inequalities can be expressed in the form $A x \ge
b$.  Note that $A$ is a matrix all of whose entries are either $-1$, 0,
or 1, and, in the first $2^N-1$ rows (the lines corresponding to
equations for each $E \subset S$), all the entries are either $0$ or
$-1$, while in the final $N+1$ rows, all the entries are either 0 or 1. 

A solution of this system of inequalities provides the desired $\Pr$.
But if this systems has no solution, then by Farkas' Lemma, there exists
a nonnegative vector $\beta$ such that $\beta A = 0$ and $\beta
b > 0$.  Since all the entries of $A$ are either $-1$, 0, or 1, it
follows from standard observations (cf. \cite[Lemma 2.7]{FHM}) we can
take $\beta$ to a vector of all whose entries are rational.%
\fullv{
\footnote{There is a slight subtlety here since $\beta$ also has to
satisfy $\beta b > 0$, and $b$ may involve irrational numbers (since
$f(E)$ may be irrational for some sets $E$).   However, if there is a
nonnegative $\beta$ that satisfies $\beta A = 0$ and $\beta b > 0$, then
there is a nonnegative $\beta$ that satisfies $\beta A = 0$ and $\beta
b' > 0$, where $b'$ consists only of rational entries and $b' \le b$.
Thus, there is a vector $\beta$ with rational entries such that $\beta A
= 0$ and $\beta b' > 0$, so $\beta b > 0$.}}  Since we can
multiply each term in $\beta$ by the product of the denominators of
the entries of $\beta$, we can assume without loss of generality that
the entries of $\beta$ are natural numbers.  

Since $A$ has $2^N + N$ rows, $\beta$ is a vector of the form
$(\beta_1, \ldots, \beta_{2^N + N})$.  Let $A_1, \ldots, A_{2^N + N}$
be the rows of $A$; each of these is a vector of length $N$.   Since 
$\beta A = 0$, that means that $\beta_1 A_1 + \cdots + \beta_{2^N+N}
A_{2^N + N} = 0$.  Suppose for now that $\beta_{2^N}$, \ldots,
$\beta_{2^N+N-1}$  (the coefficients for the rows corresponding to the
inequalities $x_i \ge 0$ for $i = 1, \ldots, N$) are all 0; as I show
below, this assumption can be made without loss of generality.

With this assumption, we can rewrite the equations as 
$\beta_1 A_1 + \ldots \beta_{2^N - 1} A_{2^N-1} = -\beta_{2^N+N}
A_{2^N+N}$.  If $E_1, \ldots, E_{2^N-1}$ are the subsets of $S$ that
correspond to the equations for $A_1, \ldots, A_{2^N-1}$, respectively, 
this equation says that
$\beta_1$ copies of $\overline{E}_1$, $\beta _2$ copies of
$\overline{E}_2$, \ldots, and $\beta_{2^N-1}$ copies of
$\overline{E}_{2^N-1} $ form a 
$\beta_{2^N+N}$-cover of $S$.  (Recall that $A_{2^N+N}$ is a row of all
$-1$'s, so $-A_{2^N+N}$ corresponds to $S$.)  
Thus, by REG3, $\beta_1 f(E_1) + \cdots + \beta_{2^N-1}
f(E_{2^N-1}) \ge \beta_{2^N+N} f(\emptyset) =  \beta_{2^N+N}$. 
But Farkas' Lemma requires that $\beta b >0$, where, by construction,
$b_i = -f(E_i)$ for $i = 1, \ldots, 2^{N} - 1$, 
$b_i = 0$ for $i = 2^N, \ldots, 2^N+N-1$, and $b_{2^N+N} = 1$.
Thus, we must have $-(\beta_1 f(E_1) + \cdots + \beta_{2^N-1}
f(E_{2^N-1}))  > -\beta_{2^N+N}$.  Clearly, this gives a
contradiction.  Thus, we can conclude, as desired, that the equations
are solvable, and that there exists a probability measure $\Pr$
such that $\alpha_{\Pr} = 1$.  

It remains to show that we can assume without loss of generality that 
$\beta^{2^N}$, \ldots, $\beta^{2^N+N-1}$ are all 0.  Note that since
$\beta \ge 0$, they must all be nonnegative.  I prove by induction on 
$\beta_{2^N} + \cdots + \beta_{2^N+N-1}$ that if there is a vector $\beta
\ge 0$ such that $\beta A = 0$ and $\beta b > 0$, then there is such a
vector with $\beta_{2^N} + \cdots + \beta_{2^N+N-1} = 0$.

So suppose that there is a solution $\beta$ 
with $\beta_{2^N} + \cdots + \beta_{2^N+N-1} > 0$.  Suppose without loss of
generality that $\beta_{2^N} > 0$.  
Recall that $A_{2^N}$ corresponds to the inequality $x_1 \ge 0$.
Choose $j \in \{0, \ldots, 2^N-1\}$ such
that $\beta_j > 0$ and $s_1  \notin E_j$.  There must be such a $j$, for
otherwise we would not have $\beta A = 0.$  Let $j'$ be such that 
$E_{j'} = E_j \union \{s_1\}$.  Define a vector
$\beta'$ such that $\beta'_{2^N} =
\beta_{2^N} - 1$, $\beta_j' = \beta_j - 1$, $\beta_{j'}' = \beta_j + 1$, and
$\beta_i' = \beta_i$ if $i \notin \{j, j', 2^N\}$.  It is easy to check
that $\beta' A = 0$ and that $\beta'_{2^N} + \cdots + \beta'_{2^N+N-1} < 
\beta_{2^N} + \cdots + \beta_{2^N+N-1}$.  It remains to show that
$\beta' b > 0$.  Since $E_j \subset E_{j'}$, we must have $f(E_j) \ge
f(E_{j'})$, so $\beta' b = \beta b + f(E_j) - f(E_{j'}) \ge \beta
b > 0$.  This completes the inductive step of the argument.

Now we must show the second required property holds, namely, that  $f(E)=
\P^+_{\regret}(E)$ for all $E \subseteq S$.  By construction, 
$\alpha_{\Pr} \Pr(\overline{E}) \le f(E)$ for all $E \subseteq S$, so it
suffices to show that there is some $\Pr \in \P$ such that 
$\alpha_{\Pr} \Pr(\overline{E}) =f(E)$.   For this, it suffices to 
show that there exists a measure $\Pr$ such that $\Pr(\overline{E}) =
1$, and for each $E' \subset S$, we have $f(E) \Pr(\overline{E'}) \le
f(E')$, since then $\alpha_{\Pr} = f(E)$, so $\alpha_{\Pr}
\Pr(\overline{E}) = f(E)$, as desired.  

To show that such a measure exists, we again construct a set of
linear inequalities much as above, and apply Farkas' Lemma.   Using the
same notation as above, suppose for simplicity that $\overline{E} =
\{s_1, \ldots, s_M\}$, where $M \le N$.  
Now the required inequalities just involve the variables $x_1, \ldots,
x_M$:
$$\begin{array}{lll}
- \sum_{\{i: s_i \in \overline{E} \inter \overline{E}'\}} x_i 
\ge -f(E')/f(E),
\shortv{\\ \qquad\qquad}
 \mbox{ for all $E' \subset S$ such that $\overline{E} \inter
\overline{E}' \ne \emptyset$}\\
x_i \ge 0, \mbox{ for $i = 1, \ldots, M$}\\
x_1 + \cdots + x_M \ge 1.
\end{array}$$
Again, the requirement that $x_1 + \cdots + x_M \le 1$ follows from the
equation for $E$.  

If this system of inequalities is satisfiable, then we have
the required probability measure, so suppose that it is not
satisfiable.  Again, writing this system of equations as $Ax \ge b$, by
Farkas' Lemma, there exists a nonnegative vector $\beta$ such that
$\beta A = 0$ and $\beta b > 0$.  We now proceed much as before.  Again,
we can assume that $\beta$ is a vector of natural numbers.
If we assume for now that $\beta^{2^M}$, \ldots,
$\beta^{2^M+M-1}$  (the coefficients for the rows corresponding to the
inequalities $x_i \ge 0$ for $i = 1, \ldots, N$) are all 0, then the fact
that $\beta A = 0$ means that we have $\beta_{2^M+M}$ cover of $E$.  
We get a contradiction to REG3 in an almost identical way to above.  
%The
%argument that we can take $\beta_{2^M}, \ldots, \beta_{2^M+M-1}$ to all
%be 0 is identical to that above.  
This completes the argument.
\eprf
}

\fullv{
\section{Conclusion}\label{conclusion}
%joe5
I have defined an approach for associating with an event $E$ a numerical
representation of its likelihood when uncertainty is represented by a
weighted set of probability measures.  
%joe7
The representation consists of a pair of a numbers, which can be thought
of as upper and lower bounds on the uncertainty.  The difference between
these numbers can be viewed as a measure of ambiguity.  
The two numbers coincide when
uncertainty is represented by a single probability.  Moreover, if each
probability measure gets weight 1, then the two numbers are essentially
be viewed as 
the lower and upper probabilities of $E$.   Thus, the approach can be
viewed as a generalization of lower and upper probability to the case of
weighted probability measures.  The definitions show that there is a
interesting connection between regret-based approaches and
minimization/maximization approaches when it comes to defining
likelihood; this
connection breaks down when it 
comes to more general utility calculations  \cite{HL12}.

The main technical result of the paper is a complete characterization of
the likelihood in the case where the state space is finite.  The notion
of likelihood can easily be extended to the case of an infinite state
space (of course, an integral has to be used instead of a sum to
calculate expected utility).  I believe that the characterization
theorem will still hold with essentially no change, although I have not
checked details carefully. 

Of course, it would be useful to get a better understanding of this
numerical representation, to see if it really captures an agent's
feelings about both the ambiguity and the risk associated with an
event, and to understand its technical properties.  I leave this to
future work.
}

\paragraph{Acknowledgments:} I thank Samantha Leung and the
reviewers of ECSQARU for many useful comments on the paper.

\fullv{\bibliographystyle{chicagor}}
\shortv{\bibliographystyle{splncs03}}
\bibliography{z,joe,riccardo}
\shortv{

}

\end{document}

%% file: defn.tex
%  THEOREM-LIKE ENVIRONMENTS

\newtheorem{THEOREM}{Theorem}[section]
\newenvironment{theorem}{\begin{THEOREM} \hspace{-.85em} {\bf :} }%
                        {\end{THEOREM}}
\newtheorem{LEMMA}[THEOREM]{Lemma}
\newenvironment{lemma}{\begin{LEMMA} \hspace{-.85em} {\bf :} }%
                      {\end{LEMMA}}
\newtheorem{COROLLARY}[THEOREM]{Corollary}
\newenvironment{corollary}{\begin{COROLLARY} \hspace{-.85em} {\bf :} }%
                          {\end{COROLLARY}}
\newtheorem{PROPOSITION}[THEOREM]{Proposition}
\newenvironment{proposition}{\begin{PROPOSITION} \hspace{-.85em} {\bf :} }%
                            {\end{PROPOSITION}}
\newtheorem{DEFINITION}[THEOREM]{Definition}
\newenvironment{definition}{\begin{DEFINITION} \hspace{-.85em} {\bf :} \rm}%
                            {\end{DEFINITION}}
\newtheorem{CLAIM}[THEOREM]{Claim}
\newenvironment{claim}{\begin{CLAIM} \hspace{-.85em} {\bf :} \rm}%
                            {\end{CLAIM}}
\newtheorem{EXAMPLE}[THEOREM]{Example}
\newenvironment{example}{\begin{EXAMPLE} \hspace{-.85em} {\bf :} \rm}%
                            {\end{EXAMPLE}}
\newtheorem{REMARK}[THEOREM]{Remark}
\newenvironment{remark}{\begin{REMARK} \hspace{-.85em} {\bf :} \rm}%
                            {\end{REMARK}}
%\newenvironment{proof}{\noindent {\bf Proof:} \hspace{.677em}}%
%                      {}

%theorem
\newcommand{\thm}{\begin{theorem}}
%lemma
\newcommand{\lem}{\begin{lemma}}
%proposition
\newcommand{\pro}{\begin{proposition}}
%definition
\newcommand{\dfn}{\begin{definition}}
%remark
\newcommand{\rem}{\begin{remark}}
%example
\newcommand{\xam}{\begin{example}}
%corollary
\newcommand{\cor}{\begin{corollary}}
%proof
\newcommand{\prf}{\noindent{\bf Proof:} }
%end theorem
\newcommand{\ethm}{\end{theorem}}
%end lemma
\newcommand{\elem}{\end{lemma}}
%end proposition
\newcommand{\epro}{\end{proposition}}
%end definition
\newcommand{\edfn}{\bbox\end{definition}}
%end remark
\newcommand{\erem}{\bbox\end{remark}}
%end example
\newcommand{\exam}{\bbox\end{example}}
%end corollary
\newcommand{\ecor}{\end{corollary}}
%end proof
\newcommand{\eprf}{\bbox\vspace{0.1in}}
%begin equation
\newcommand{\beqn}{\begin{equation}}
%end equation
\newcommand{\eeqn}{\end{equation}}
% white box

%black box
\newcommand{\bbox}{\vrule height7pt width4pt depth1pt}

\newcommand{\clm}{\begin{claim}}
\newcommand{\eclm}{\end{claim}}
% (not)member of

% \sub will be used for subscript.

% \su will be used for superscript.

%right arrow

%left arrow

%bold face lower-case letters
%for bold Greek symbols in math mode (with \boldsymbol{\sigma}, etc.)
%\newcommand{\boldsymbol}[1]{\mbox{\boldmath $\bf #1$}}

%bold face upper-case letters

%double turnstile

%single turnstile

%fat right arrow

%fat left arrow

%fat double arrow

%big or

%big and

%union
\newcommand{\union}{\cup}
%intersection
\newcommand{\inter}{\cap}
%bold letters

%\newfont{\sqi}{cmssqi8}

%Use $\IC \;\;$
% multivalued arrow

% phi
\renewcommand{\phi}{\varphi}
%\renewcommand{\Diamond}{{\bf Large \diamond}}

%binomial coefficient:

%\newcommand{\fullv}[1]{#1}
%\newcommand{\shortv}{\commentout}

% Joe's Section

%\H, \L, \O, \P and \S already taken; but we're redefining \P anyway

%\newcommand{\H}{{\cal H}}

%\newcommand{\L}{{\cal L}}

%\newcommand{\O}{{\cal O}}

\renewcommand{\P}{{\cal P}}

%\newcommand{\S}{{\cal S}}

 %vertical bar with space around it
 %colon with space around it

%\renewcommand{\Box}{\mathbin{\vcenter{\hrule
%    \hbox{\vrule \kern .6em
%          \vbox to .6em{}\vrule}\hrule}}\hspace{.17ex}}

\newcommand{\ol}{\setlength{\itemsep}{0pt}\begin{enumerate}}
\newcommand{\eol}{\end{enumerate}\setlength{\itemsep}{-\parsep}}
\newcommand{\ul}{\setlength{\itemsep}{0pt}\begin{itemize}}
\newcommand{\dl}{\setlength{\itemsep}{0pt}\begin{description}}
\newcommand{\edl}{\end{description}\setlength{\itemsep}{-\parsep}}
\newcommand{\eul}{\end{itemize}\setlength{\itemsep}{-\parsep}}

%chck macros

%\newcommand{\IcR}{{\cal I}_{\cal R}}
%\newcommand{\IRca}{{\cal I}_\Rca}
%\newcommand{\Rca}{{\cR_{ca}}}

%\newcommand{\RP}{{\cR_P}}
%\newcommand{\IRP}{{\cal I}_\RP}
%\newcommand{\RSBA}{{\cR_{\it sba}}}
%\newcommand{\IRSBA}{{\cal I}_\RSBA}

%\newcommand{\RF}{\cR_{\scriptscriptstyle {\cal F}}}
%\newcommand\eqdef{\buildrel {\rm def}\over =}

%book macros

%chguide macros

%\newcommand{\MPrt}{{\cal M}_n^{rt}}

%\newcommand{\MPrst}{{\cal M}_n^{rst}}

%\newcommand{\MPelt}{{\cal M}_n^{elt}}

%\newcommand{\CSn}{\I_n^{cs}(\Phi)}
%\newcommand{\CSn}{\I_n^{oa}(\Phi)}
%\newcommand{\CSnm}{\I_n^{oa}}
%\newcommand{\CSnp}{\I_n^{cs}(\Phi')}

%\newcommand{\CSc}{\C_n^{cs}(\Phi)}

%\newcommand{\Ccs}{\C_n^{cs}}

%ron
%\newcommand{\CSAX}{CS$_{{{n}},\Phi}$}
%\newcommand{\CSAXN}{CS$_{{{n}},\Phi}'$}

%\newcommand{\IKB}{\I_n^{KB}}

\newcommand{\commentout}[1]{}

\newcommand{\bi}{\begin{itemize}}
\newcommand{\ei}{\end{itemize}}
\newcommand{\be}{\begin{enumerate}}
\newcommand{\ee}{\end{enumerate}}

%% file: spage.tex
\setlength{\evensidemargin}{0in}
\setlength{\oddsidemargin}{0in}
\setlength{\textwidth}{6.25in}
\setlength{\textheight}{8.5in}
\setlength{\topmargin}{0in}
\setlength{\headheight}{0in}
\setlength{\headsep}{0in}
\setlength{\itemsep}{0pt}

\setlength{\parskip}{\smallskipamount}